\documentclass[11pt,a4paper]{article}

\usepackage{amssymb}  
\usepackage{amsmath}  
\usepackage{pifont}   
\usepackage{graphicx} 
\usepackage{enumitem} 
\usepackage{changepage} 
\usepackage{wrapfig} 

\usepackage[english]{babel} 
\usepackage{mathptmx} 
\usepackage{authblk} 
\usepackage{fullpage} 
\usepackage[margin=10pt,labelfont=bf]{caption} 
\usepackage[colorlinks=true,citecolor=black,linkcolor=black,urlcolor=black]{hyperref} 
\usepackage[bottom]{footmisc} 

\graphicspath{{figures/}}

\newcommand{\figref}[1]{Fig.~\ref{#1}}
\newcommand{\tabref}[1]{Table~\ref{#1}}

\author[2]{
Claudio Michaelis\textsuperscript{1,$\S$},
Matthias Bethge\textsuperscript{1} \&
Alexander S. Ecker\textsuperscript{}}

\affil[1]{University of Tübingen, Germany}
\affil[2]{University of Göttingen, Germany}
\affil[$\S$]{ \texttt{claudio.michaelis@uni-tuebingen.de}}

\title{A Broad Dataset is All You Need for One-Shot Object Detection}

\begin{document}

\maketitle

\begin{abstract}
    Is it possible to detect arbitrary objects from a single example? A central problem of all existing attempts at one-shot object detection is the generalization gap: Object categories used during training are detected much more reliably than novel ones. We here show that this generalization gap can be nearly closed by increasing the number of object categories used during training. Doing so allows us to improve generalization from seen to unseen classes from 45\% to 89\% and improve the state-of-the-art on COCO by 5.4 \%AP\textsuperscript{50} (from 22.0 to 27.5). 
    We verify that the effect is caused by the number of categories and not the number of training samples, and that it holds for different models, backbones and datasets. This result suggests that the key to strong few-shot detection models may not lie in sophisticated metric learning approaches, but instead simply in scaling the number of categories. We hope that our findings will help to better understand the challenges of few-shot learning and encourage future data annotation efforts to focus on wider datasets with a broader set of categories rather than gathering more samples per category.
\end{abstract}



\begin{figure}[ht]
    \centering
    \includegraphics[width=\linewidth]{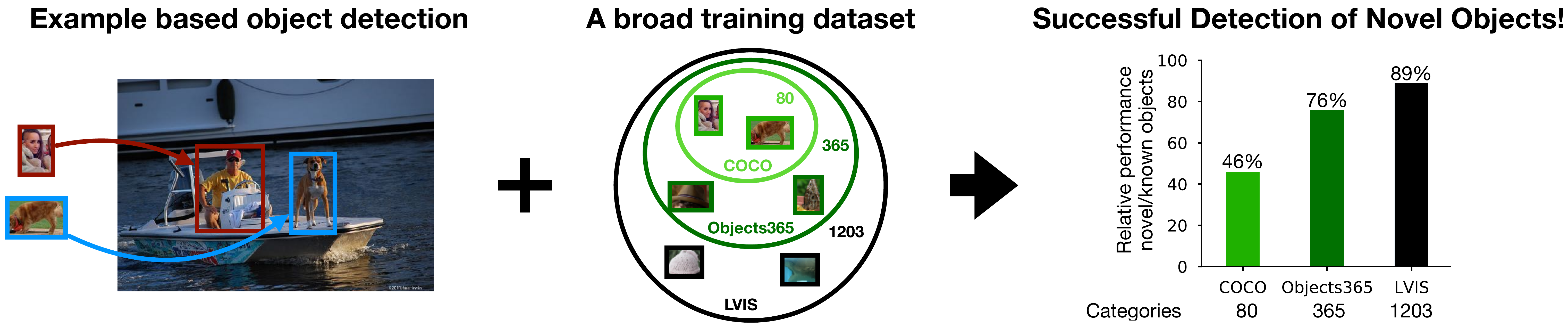}
    \caption{Example based object detectors can in theory detect any object based on an example image. However existing models trained on the datasets with few categories such as COCO perform significantly worse for novel than known objects. We here show that this generalization gap progressively shrinks when training the same models with more categories thus moving us closer to models which can actually detect any object.}
    \label{fig:teaser}
\end{figure}

\section{Introduction}


\begin{adjustwidth}{20pt}{20pt}
\textit{It's January 2021 and your long awaited household robot finally arrives. Equipped with the latest ``Deep Learning Technology'', it can recognize over 21,000 objects. Your initial excitement quickly vanishes as you realize that your casserole is not one of them. When you contact customer service they ask you to send some pictures of the casserole so they can fix this. They tell you that the fix will be some time, though, as they need to collect about a thousand images of casseroles to retrain the neural network. While you are making the call your robot knocks over the olive oil because the steam coming from the pot of boiling water confused it. You start filling out the return form ... }\\
\end{adjustwidth}

While not 100\% realistic, the above story highlights an important obstacle towards truly autonomous agents: such systems should be able to detect novel, previously unseen objects and learn to recognize them based on (ideally) a single example. Solving this one-shot object detection problem can be decomposed into three subproblems: (1) designing a class-agnostic object proposal mechanism that detects both known and previously unseen objects; (2) learning a suitably general visual representation (metric) that supports recognition of the detected objects; (3) continuously updating the classifier to accommodate new object classes or training examples of existing classes. In this paper, we focus on the detection and representation learning part of the pipeline, and we ask: what does it take to learn a visual representation that allows detection and recognition of previously unseen object categories based on a single example?

We operationalize this question using an example-based visual search task (\figref{fig:teaser}) that has been investigated before using handwritten characters (Omniglot; \cite{michaelis2018clutter}) and real-world image datasets (Pascal VOC, COCO; \cite{michaelis2018instance, Hsieh2019coae, Zhang2019comparison, Fan2020fsod, Li2020woft}). Our central hypothesis is that scaling up the number of object categories used for training should improve the generalization capabilities of the learned representation. This hypothesis is motivated by the following observations. On (cluttered) Omniglot~\cite{michaelis2018clutter}, recognition of novel characters works almost as well as for characters seen during training. In this case, sampling enough categories during training relative to the visual complexity of the objects is sufficient to learn a metric that generalizes to novel categories. In contrast, models trained on visually more complex datasets like Pascal VOC and COCO exhibit a large generalization gap: novel categories are detected much less reliably than ones seen during training. This result suggests that on the natural image datasets, the number of categories is too small given the visual complexity of the objects and the models retreat to a shortcut \cite{geirhos2020shortcut} -- memorizing the training categories. 

To test the hypothesis that wider datasets improve generalization, we increase the number of object categories during training by using datasets (LVIS, Objects365) that have a larger number of categories annotated. Our experiments support this hypothesis and suggest the following conclusions:
\begin{itemize}
    \item The generalization gap between training and novel categories is a key problem in one-shot object detection.
    \item This generalization gap can be almost closed by increasing the number of categories used for training: going from 80 classses in COCO to 1200 in LVIS improves relative performance from 45\% to 89\%.
    \item A detailed analysis shows that number of categories, not the amount of data, is the driving force behind this effect.
    \item Closing the generalization gap allows using established methods from the object detection community (like e.g. stronger backbones) to improve performance on known and novel categories alike. 
    \item We use these insights to improve state-of-the-art performance on COCO by \textbf{5.4} \%AP\textsuperscript{50} (from 22 \%AP\textsuperscript{50} to 27.5 \%AP\textsuperscript{50}) using annotations from LVIS.
\end{itemize}

\section{Related Work}

\textbf{Object detection$\quad$}
Object detection - the task of detecting objects in complex, cluttered scenes - has seen huge progress since the widespread adoption of DNNs \cite{Girshick2014rcnn, Ren2015faster, He2017mask, Lin2017fpn, Chen2019mmdetection, Wu2019detectron2, Carion2020detr}. Similarly the number of datasets has grown steadily, fueled by the importance this task has for computer vision applications \cite{Everingham2010pascal, imagenet, Lin2014coco, Zhou2017ade, mvd2017, OpenImages, Gupta2019lvis,  Shao2019objects365}. However most models and datasets focus on scenarios where abundant examples per category are available.

\textbf{Few-shot learning$\quad$}
Algorithms for few-shot learning - learning a model from only a few examples - can broadly be separated into two categories: Metric learning\cite{Koch2015siamese, Vinyals2016matching, Snell2017proto} - learn a good representation and metric that generalizes to new data. And meta learning \cite{Finn2017maml, rusu2018leo} - learn a good way to learn a new task. However, recent work has shown that complex algorithmic approaches can be rivaled by improving and scaling simple approaches like transfer learning \cite{chen2019closer, nakamura2019revisiting, dhillon2019baseline, kolesnikov2019bigtransfer}.

\textbf{Few-shot \& one-shot object detection$\quad$}
Recently, several groups have started to tackle few-shot learning for object detection. Two training and evaluation paradigms have emerged. The first is inspired by continual learning: incorporate a set of new categories with only a few labeled images per category into an existing classifier \cite{kang2018few, Yan2019metar-cnn, Wang2019meta, Wang2020frustratingly}. The second one phrases the problem as an example-based visual search: detect objects based on a single example image \cite[\figref{fig:teaser} left]{michaelis2018instance, Hsieh2019coae, Zhang2019comparison, Fan2020fsod, Li2020woft}. We refer to the former (continual learning) as \emph{few-shot object detection}, since typically 10--30 images are used for experiments on COCO. In contrast, we refer to the latter (visual search) as \emph{one-shot object detection}, since the focus is on the setting with a single example. In the present paper we work with this latter paradigm, since it focuses on the representation learning part of the problem and avoids the additional complexity of continual learning.


\textbf{Methods for one-shot object detection$\quad$}
Existing methods for one-shot object detection usually combine a standard object detection architecture with metric or meta-learning methods \cite{Biswas2015laplacian, michaelis2018instance, Hsieh2019coae, Zhang2019comparison, Fan2020fsod, Osokin2020os2d, Li2020woft}. To better handle complex scenes and pose changes methods such as spatial awareness \cite{Li2020woft} or pose transforms \cite{Biswas2015laplacian, Osokin2020os2d} have been proposed. A recent method uses a transformer to solve the matching problem \cite{chen2021ait}. We here use one of the most straightforward models, Siamese Faster R-CNN \cite{michaelis2018instance}, to demonstrate that a change of the training data rather than the model architecture is sufficient to substantially reduce the generalization gap between known and novel categories.

\textbf{Number of categories in few-shot learning$\quad$}
Most of the few-shot learning literature focuses on developing algorithmic solutions to a set of existing small-scale benchmarks. In contrast a lot less attention has been payed to exploring new tasks or datasets. The influence of the training data was mostly observed indirectly, e.g. through better performance on datasets with more categories such as \textit{tiered}ImageNet vs. \textit{mini}ImageNet. We here flip the focus demonstrating that significant progress can be made by keeping the algorithm the same and only changing the training data. Concurrent studies confirm this finding that more categories help few-shot object detection \cite{Fan2020fsod} and few-shot image classification \cite{sbai2020fewshotdata, Jiang2020fewshotdata}. We add to this by not only looking at few-shot performance but comparing it with performance on known categories (generalization gap). This allows us to uncover the functional relationship behind the effect (closing a shortcut). 


\section{Experiments}

\begin{figure}[b]
    \centering
    \includegraphics[width=0.7\linewidth]{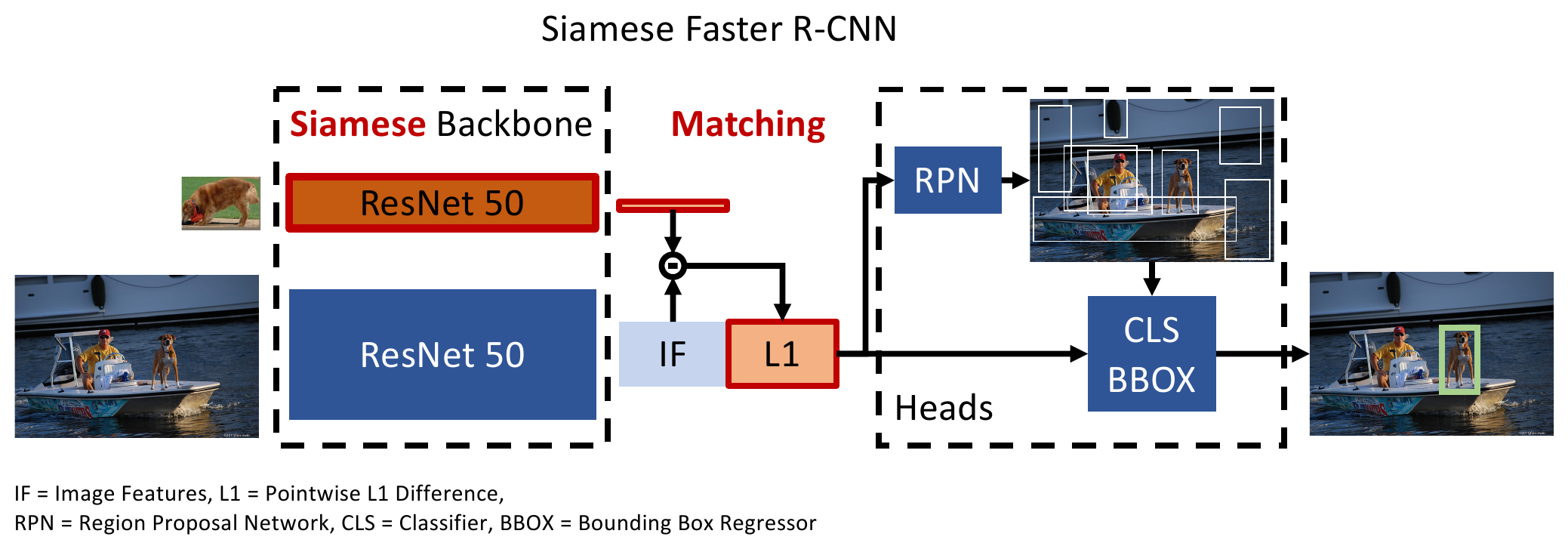}
    \caption{Siamese Faster R-CNN}
    \label{fig:siamese_frcnn}
\end{figure}

\textbf{Models$\quad$}
We mainly use Siamese Faster R-CNN, an example-based version of Faster R-CNN \cite{Ren2015faster} similar to Siamese Mask R-CNN \cite{michaelis2018instance}. Briefly, it consists of a feature extractor, a matching step and a standard region proposal network and bounding box head (\figref{fig:siamese_frcnn}). The feature extractor (called backbone in object detection) is a standard ResNet-50 with feature pyramid networks \cite{He2016resnet, Lin2017fpn} which is applied to the image and reference with weight sharing. In the matching step the reference representation is compared to the image representation in a sliding window approach by computing a feature-wise L1 difference. The resulting similarity encoding representation is concatenated to the image representation and passed on to the region proposal network (RPN). The RPN proposes a set of bounding boxes which potentially contain objects. These boxes are then classified as containing an object from the reference class or something else (other object or background). Box coordinates are refined by bounding box regression and overlapping boxes are removed using non-maximum suppression.

We additionally developed Siamese RetinaNet, a single-stage detector based on RetinaNet \cite{Lin2017retinanet}. The feature extraction and matching steps are identical to Siamese Faster R-CNN, but it uses the unified RetinaHead to jointly propose and classify bounding boxes. To counter the effect of too many negative samples, the classifier is trained with focal loss \cite{Lin2017retinanet}.

\textbf{Training \& Evaluation$\quad$}
The example-based training is slightly different from the traditional object detection training paradigm. For each image a reference category is randomly chosen by picking a category with at least one instance in the image. A reference is retrieved by randomly selecting one instance from this category in another image and tightly cropping it.
The labels for each bounding box are changed to 0 or 1 depending on whether the object is from the reference category or not. Annotations for objects from the held-out categories are removed from the dataset before training. At test time a similar procedure is chosen but instead of picking one category for each image, all categories with at least one object in the image are chosen \cite{michaelis2018instance} and one (1-shot) or five (5-shot) reference images are provided. Predictions are assigned their corresponding category label and evaluation is performed using standard tools and metrics.

\textbf{Implementation$\quad$}
We implemented Siamese Faster R-CNN and Siamese RetinaNet in mmdetection v1.0rc \cite{Chen2019mmdetection}, which improved performance by more than 30\% over the original Siamese Mask R-CNN \cite[\tabref{table:coco_sota}]{michaelis2018instance}. We keep all hyperparameters the same as in the standard Faster R-CNN implementation of mmdetection. Due to resource constraints we reduce the number of samples per epoch to 120k for Objects365.

\textbf{Hyperparameters$\quad$}
Our model is derived from mmdetection v1.0rc \cite{Chen2019mmdetection} and uses the same hyperparameters as used for Faster R-CNN and RetinaNet\footnote{All details can be found in the respective configs: \url{https://github.com/open-mmlab/mmdetection/tree/5bf935e1b7621b234ddb34ef6c32b2b524243995/configs}}.
Please note that the default settings for Pascal VOC differ slightly from those for COCO training. We use the COCO hyperparameters for experiments on COCO, LVIS and Objects365 and Pascal VOC settings for Pascal VOC.


\begin{table}[b]
    \centering
    \begin{small}
    \begin{tabular}{lc|rrrrrc}
Dataset    & Version & Classes &  Images & Instances & Ins/Img & Cls/Img & Thr.\\
\hline\hline
Pascal VOC & 07+12 &       20 &       8k &      23k     & 2.9 &     1.6 & \checkmark\\
COCO       & 2017  &       80 &     118k &     860k    & 7.3 &     2.9 & \checkmark\\
LVIS       & v1    &     1,203 &      100k &     1.27M  & $\geq$12.8$^*$\hspace{-4.4pt}  & $\geq$3.6$^*$\hspace{-4.4pt} & \ding{55}\\
Objects365 & v2    &      365 &       1.94M &     28M & 14.6 &     6.1 & \checkmark\\
\hline
    \end{tabular}
    \caption{Dataset comparison. Thr. = Throughoutly annotated: every instance of every class is annotated in every image. $^*$LVIS has potentially more objects and categories per image than are annotated due to the non-exhaustive labeling.}
    \label{table:datasets}
    \end{small}
\end{table}

\textbf{Datasets$\quad$}
We use the four datasets shown in \tabref{table:datasets}: COCO \cite{Lin2014coco}, Objects365 \cite{Shao2019objects365}, LVIS \cite{Gupta2019lvis} and Pascal VOC \cite{Everingham2010pascal}. We use standard splits and test on the validation sets except for Pascal VOC where we test on the 2007 test set. Due to resource constraints, we evaluate Objects365 on a fixed subset of 10k images from the validation set. 

\textbf{Category splits$\quad$}
Following common protocol for example-based detection \cite{michaelis2018instance, Shaban2017oneshot} we split the categories in each dataset into four splits using every fourth category as hold-out set and the other 3/4 categories for training. So on Pascal VOC there are 15 categories for training in each split, on COCO there are 60, on Objects365 274 and on LVIS 902. We train and test four models (one for each split) and report the mean over those four models, so performance is always measured on all categories. Computing performance in this way across all categories is preferable to using a fixed subset as some categories may be harder than others. During evaluation, the reference images are chosen randomly. We therefore run the evaluation five times, reporting the average $\text{AP}^{50}$ over splits. The 95\% confidence intervals for the average $\text{AP}^{50}$ is below $\pm0.2\%\text{AP}^{50}$ for all experiments.

\begin{figure}[t]
    \centering
    \includegraphics[width=.94\linewidth]{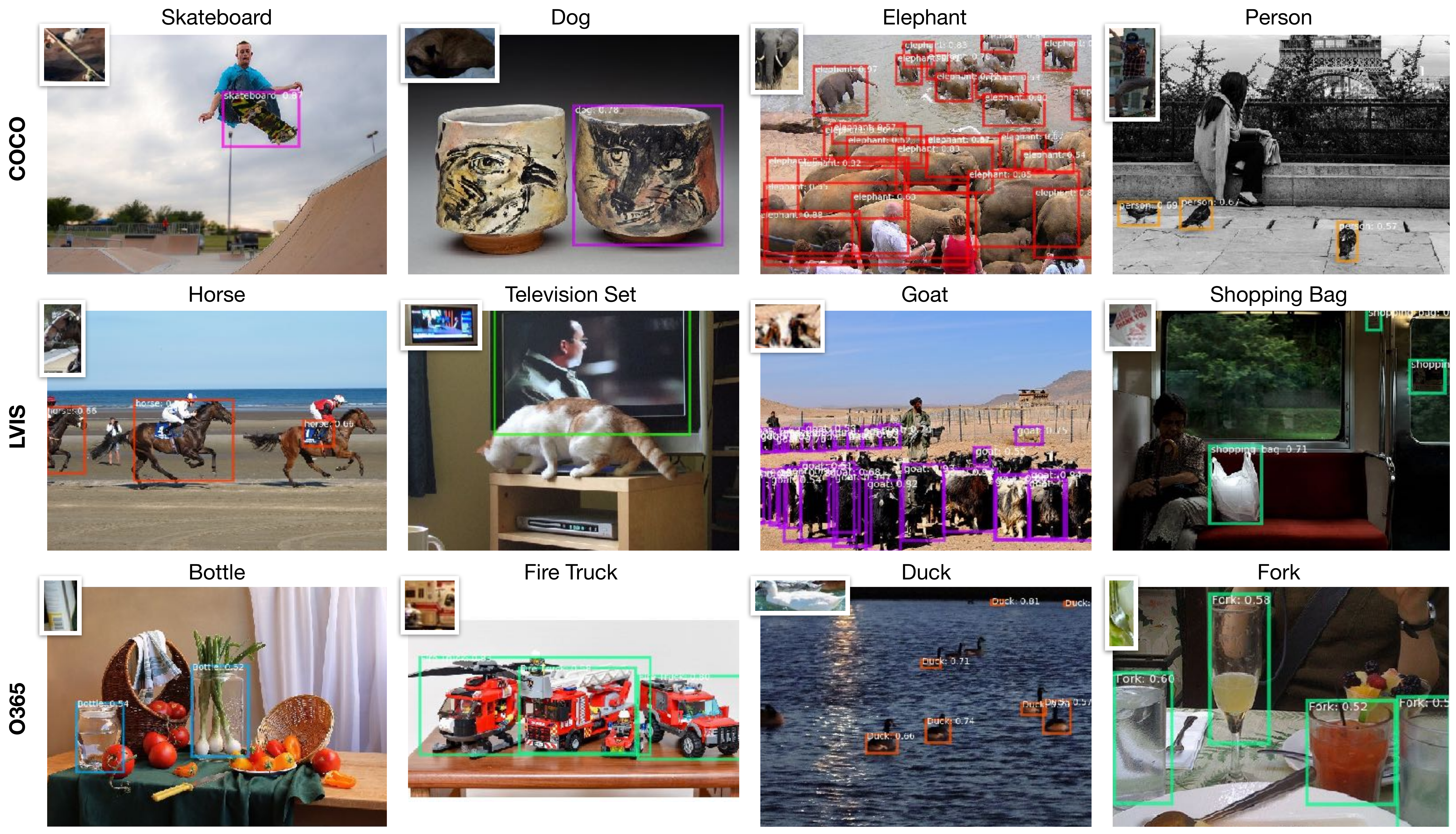}
    \caption{Example predictions on held-out categories (ResNet-50 backbone). The left three columns show success cases. 
    The rightmost column shows failure cases in which objects are overlooked and/or wrongfully detected.}
    \label{fig:examples}
\end{figure}

\section{Results}

\subsection{Generalization gap on COCO and Pascal VOC}

\begin{wrapfigure}[12]{r}{0.28\textwidth}
    \vspace{-12pt}
    \includegraphics[width=\linewidth]{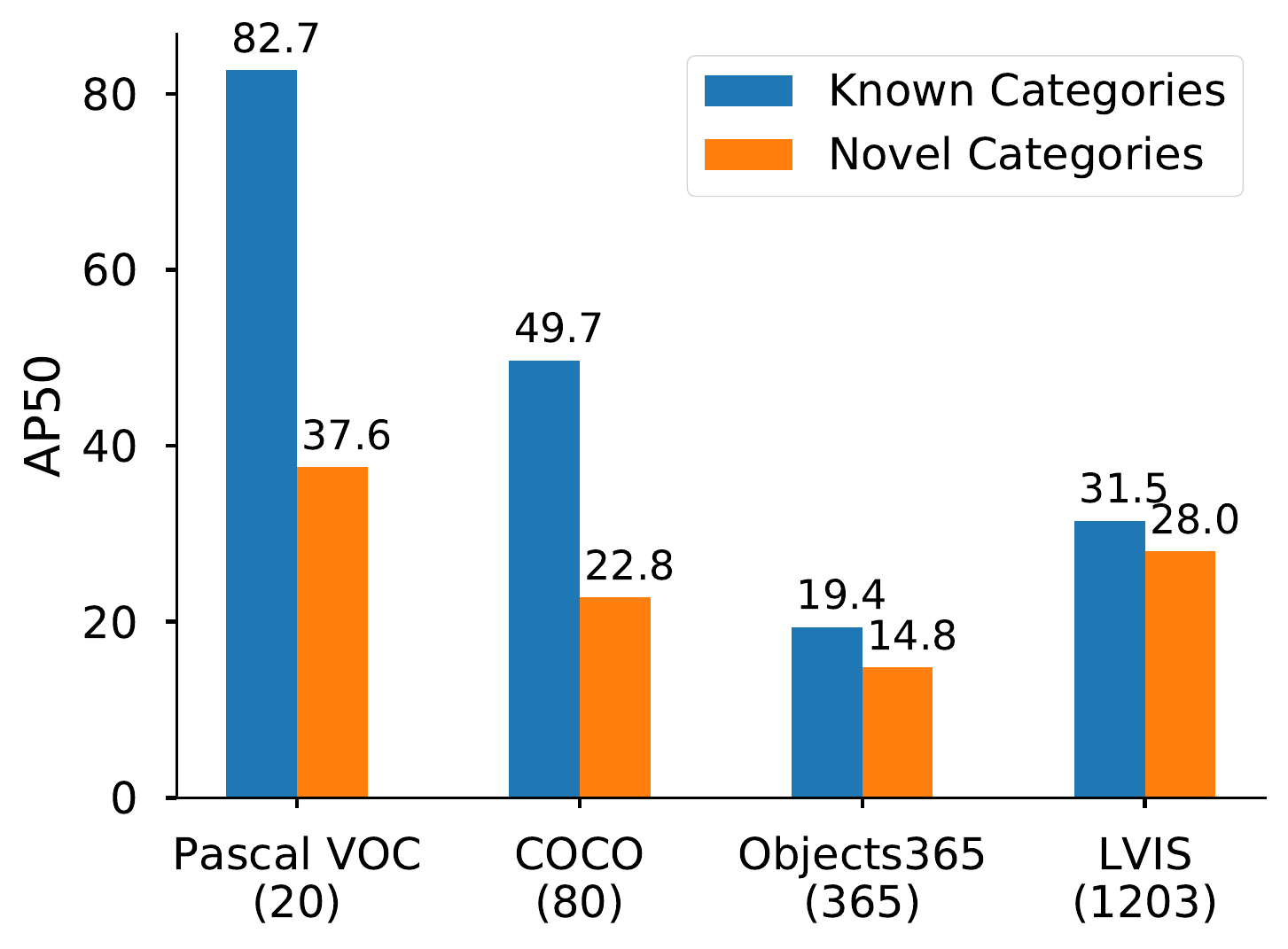}
    \caption{Performance on known and novel categories for different datasets.}
    \label{fig:absolute_performance}
\end{wrapfigure}

We start by showing that objects of held-out categories are detected less reliably on COCO and Pascal VOC. On both datasets, Siamese Faster R-CNN shows strong signs of overfitting to the training categories (\figref{fig:absolute_performance} \& \tabref{table:baselines}): despite setting a new stat-of-the-art performance is much higher than for categories held-out during training (COCO: $49.7 \rightarrow 22.8 \ \% \text{AP}^{50}$; Pascal VOC: $82.7 \rightarrow 37.6 \ \% \text{AP}^{50}$). We refer to this drop in performance as the \emph{generalization gap}. This result is consistent with the literature: 
\cite{Hsieh2019coae} --~the previous state-of-the-art~-- report performance dropping $40.9 \rightarrow 22.0 \ \% \text{AP}^{50}$ on COCO (see \tabref{table:coco_sota} below). Some newer models reportedly close the gap on Pascal VOC \cite{Zhang2019comparison, Hsieh2019coae, Li2020woft}; we will discuss Pascal VOC further in the next section. Example predictions show good localization (bounding boxes) even for unknown objects in cluttered scenes while classification errors make up the majority of mistakes (\figref{fig:examples}).

\subsection{Pascal VOC is too easy to evaluate one-shot object detection models}

Having identified this large generalization gap, we ask whether the models have learned a useful metric for one-shot detection at all or whether they rely on simple dataset statistics.
Pascal VOC contains, on average, only 1.6 categories and 2.9 instances per image. In this case, simply detecting all foreground objects may be a viable strategy. To test how well such a trivial strategy would perform, we provide the model with uninformative references (we use all-black images). Interestingly, this baseline performs very well, achieving $59.6 \ \% \text{AP}^{50}$ on training and $33.2 \ \% \text{AP}^{50}$ on held-out categories (\tabref{table:baselines}). For held-out categories, the difference to an example-based search is marginal ($33.2 \rightarrow 37.6 \ \% \text{AP}^{50}$). This result demonstrates that on Pascal VOC the model mostly follows a shortcut and uses basic dataset statistics to solve the task.

In contrast, COCO represents a drastic increase in image complexity compared with Pascal VOC: it contains, on average, 2.9 categories and 7.3 instances per image. As expected, in this case the trivial baseline with uninformative references performs substantially worse than the example-based search (training: $49.7 \rightarrow 10.1 \ \% \text{AP}^{50}$; held-out: $22.8 \rightarrow 4.4 \ \% \text{AP}^{50}$; \tabref{table:baselines}). Thus, the added image complexity in COCO forces the model to use the reference image for classification but the small set of categories is not sufficient to prevent memorizing the training categories.

\begin{table}[t]
    \centering
    \begin{footnotesize}
    \begin{tabular}{l|cc|cc}
    & \multicolumn{2}{c}{COCO} & \multicolumn{2}{c}{Pascal VOC}\\
    Categories$\rightarrow$ & Train & Held-Out & Train & Held-Out\\
    \hline \hline 
    Siam. Faster R-CNN & 49.7 & 22.8 & 82.7 &  37.6\\
    --- empty Refs. & 10.1 & 4.4 & 59.6 & 33.2\\
    \hline
    \end{tabular}
    \end{footnotesize}
    \caption{On COCO and Pascal VOC there is a clear performance gap (AP\textsuperscript{50}) between categories used during training (Train) and held-out categories (Held-Out). A baseline getting a black image as reference which contains no information about the target category (-- empty Refs.) performs surprisingly well on Pascal VOC but fails on COCO.}
    \label{table:baselines}
\end{table}

\subsection{Training on more categories reduces the generalization gap}

We now turn to our main hypothesis that increasing the number of categories used during training could close the generalization gap identified above. To this end we use Objects365 and LVIS, two fairly new datasets with 365 and 1203 categories, respectively (much more than the 20/80 in Pascal VOC/COCO). Indeed, training on these wider datasets improves the relative performance on the held-out categories from 46\% on COCO to 76\% on Objects365 and up to 89\% on LVIS (\figref{fig:teaser}). In absolute numbers this means going from a $26.9 \ \% \text{AP}^{50}$ gap on COCO to a $4.6 \ \% \text{AP}^{50}$ gap on Objects365 and a $3.5 \ \% \text{AP}^{50}$ gap on LVIS (\tabref{table:bigger_models}) in the one-shot setting. Increasing the number of references to five (5-shot) improves performance on all datasets but leaves relative performance unchanged (\tabref{table:bigger_models}, right columns).

\begin{figure}[b]
    \centering
    \includegraphics[width=1.00\linewidth]{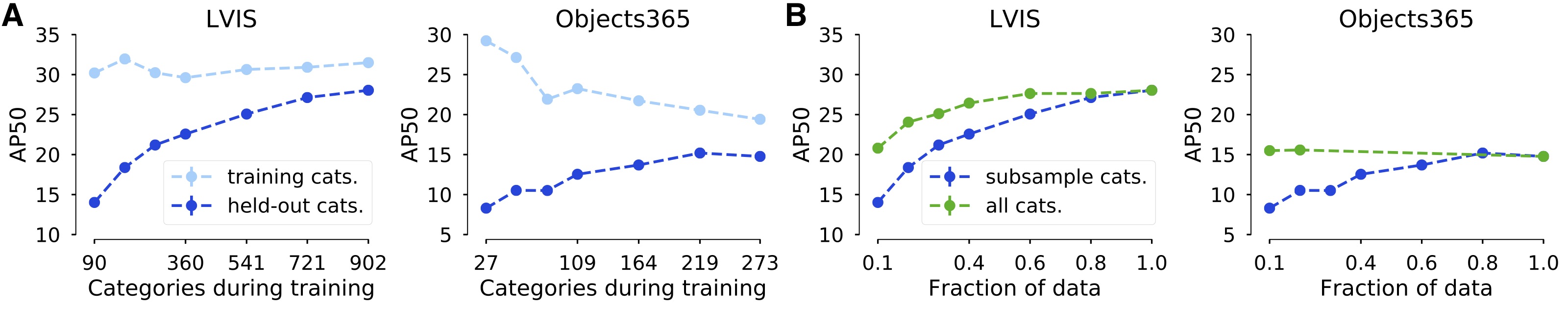}
    \caption{\textbf{A.} Experiment subsampling LVIS and Objects365 categories during training. When more categories are used during training performance on held-out categories (blue) improves while performance on the training categories (light blue) stays flat or decreases. \textbf{B.}
    Comparison of the performance on held-out categories if a fixed number of instances is chosen either from all categories (green) or from a subset of categories (blue). Having more categories is more important than having more samples per category. (1-shot results, for 5-shot see ~\figref{fig:class_fraction_5-shot})
    }
    \label{fig:class_fraction}
\end{figure}

This effect is not caused simply by differences between the datasets, as the following experiment shows: For both datasets (LVIS and Objects365), we train models on progressively more categories. When training on less than 100 categories (resembling training on COCO), a clear generalization gap is visible on both LVIS and Objects365 (\figref{fig:class_fraction}A: leftmost data points). Increasing the number of training categories leads to better performance on the held-out categories, while performance on the training categories stays the same (LVIS) or decreases (Objects365). The effect is the same in the 5-shot setting but with a better baseline performance (\figref{fig:class_fraction_5-shot} in Appendix).

\subsection{The number of categories is the crucial factor}

The results so far show that increasing the number of categories used during training reduces the generalization gap and improves performance. However, this effect could also be caused by the fact that with more categories there is also more data available. Consider the situation where we train on 10\% of the categories (90 in the case of LVIS). As we sample these categories uniformly from the dataset, we use only approximately 10\% of the total number of instances. To control for this confound, we created training sets that match the number of instances: in this case we use only 10\% of the instances in the dataset but sample them uniformly from all 900 training categories. 

The results can be seen in \figref{fig:class_fraction}B. Our example from above with 10\% of the data corresponds to the leftmost datapoint in both plots. The model trained with more categories (green) clearly outperforms the model with more instances per category (blue). The same performance gap can be seen for any fraction of the data. Thus, for a given budget of instances (labels) it is better to cover more categories than to collect as many samples per category as possible. 

\subsection{Once the generalization gap is closed more powerful models benefit novel categories}

\begin{table}[t]
    \centering
    \begin{footnotesize}
    \begin{tabular}{l|cc|ccc|ccc}
    \multicolumn{9}{c}{\textbf{COCO}} \\
    \multicolumn{3}{c|}{} & \multicolumn{3}{c|}{1-shot} & \multicolumn{3}{c}{5-shot} \\
    Model & Backb. & Sched. & Train C. & Held-Out C. & Delta & Train C. & Held-Out C. & Delta \\
    \hline \hline 
    Siam. RetinaNet & R50 & 1x & 50.6 & 18.9 & 31.7 & 55.5 & 22.1 & 33.4\\
    Siam. FRCNN & R50 & 1x & 49.7 & 22.8 & 26.9 & 54.9 & 27.6 & 27.3\\
    Siam. FRCNN & R50 & 3x & 51.7 & 21.9 & 29.8 & 57.6 & 26.7 & 30.9\\
    Siam. FRCNN & X101 & 1x & 56.4 & 23.5 & 32.9 & 61.9 & 28.6 & 33.3\\
    \hline
    \multicolumn{9}{c}{} \\
    \multicolumn{9}{c}{\textbf{LVIS}} \\
    \multicolumn{3}{c|}{} & \multicolumn{3}{c|}{1-shot} & \multicolumn{3}{c}{5-shot} \\
    Model & Backb. & Sched. & Train C. & Held-Out C. & Delta & Train C. & Held-Out C. & Delta \\
    \hline \hline 
    Siam. RetinaNet & R50 & 1x & 28.4 & 24.7 & 3.7 & 31.6 & 27.5 & 4.1\\
    Siam. FRCNN & R50 & 1x & 31.5 & 28.0 & 3.5 & 37.0 & 33.0 & 4.0\\
    Siam. FRCNN & R50 & 3x & 32.7 & 28.7 & 4.0  & 38.2 & 33.5 & 4.7\\
    Siam. FRCNN & X101 & 1x & 35.4 & 31.3 & 4.1 & 41.4 & 36.3 & 5.1\\
    \hline
    \multicolumn{9}{c}{} \\
    \multicolumn{9}{c}{\textbf{Objects365}} \\
    \multicolumn{3}{c|}{} & \multicolumn{3}{c|}{1-shot} & \multicolumn{3}{c}{5-shot} \\
    Model & Backb. & Sched. & Train Cats. & Held-Out C. & Delta & Train C. & Held-Out C. & Delta \\
    \hline \hline 
    Siam. RetinaNet & R50 & 1x & 19.7 & 14.5 & 5.2 & 23.4 & 17.2 & 6.2\\
    Siam. FRCNN & R50 & 1x & 19.4 & 14.8 & 4.6 & 25.7 & 19.9 & 5.8\\
    Siam. FRCNN & R50 & 3x & 22.0 & 16.5 & 5.5 & 27.7 & 20.9 & 6.8\\
    Siam. FRCNN & X101 & 1x & 25.0 & 17.9 & 7.1 & 30.6 & 22.4 & 8.2\\
    \hline
    \end{tabular}
    \end{footnotesize}
    \caption{Effect of a three times longer training schedule and a larger backbone (ResNeXt-101 32x4d) on model performance across datasets. While larger models and longer training times lead to no or only minor improvements on held-out categories on COCO, they do have a larger effect on LVIS and Objects365.}
    \label{table:bigger_models}
\end{table}

If models indeed learn the distribution over categories, stronger models that can learn more powerful representations should perform better on known and novel categories alike. We test this hypothesis in two ways: first, by replacing the standard ResNet-50 \cite{He2016resnet} backbone with a more expressive ResNeXt-101 \cite{xie2017resnext}; second, by using a three times longer training schedule. 

The larger backbone does not improve performance on the held-out categories on COCO (\tabref{table:bigger_models}). Instead the additional capacity is used to memorize the training categories, which is evident from the large improvement ($6.7\,\% \text{AP}^{50}$) in performance on the training categories, but only a small improvement ($0.7\,\% \text{AP}^{50}$) on the held-out categories.
In contrast, on LVIS and Objects365 the gains of the bigger backbone are not confined to the training categories but apply to the one-shot setting as well. Only a small difference remains on Objects365 ($3.0\,\% \text{AP}^{50}$ vs. $1.4\,\% \text{AP}^{50}$).

Longer training schedules show the same pattern. For COCO, performance on the training categories improves while performance on held-out categories even gets a bit worse on a 3x schedule (\tabref{table:bigger_models}). In contrast, performance on LVIS and Objects365 improves for both training and held-out categories alike, suggesting that the models do not overfit only the training categories.

\subsection{Results hold for different model configurations}

To test if our findings apply to single-stage detectors as well, we train and test Siamese RetinaNet on COCO, LVIS and Objects365 (\tabref{table:bigger_models}). Results are very similar to Siamese Faster R-CNN. Siamese RetinaNet shows a slightly larger generalization gap on COCO (relative performance: Retina: 37\% vs. FRCNN: 46\%) but results are very similar on LVIS (Retina: 87\% vs. FRCNN: 89\%) and Objects365 (Retina: 74\% vs. FRCNN: 76\%). 

Taken together we observe the same patterns for single- and two-stage detectors with different backbones and learning rate schedules on two datasets (Objects365 and LVIS) for 1-shot and 5-shot evaluation. This suggests that our conclusions may extend to most object detection models and we can expect to significantly boost performance using the large toolboxes which exist for traditional object detection.

\subsection{State-of-the-art on COCO using LVIS}

\begin{figure}[t]
    \centering
    \includegraphics[width=0.98\linewidth]{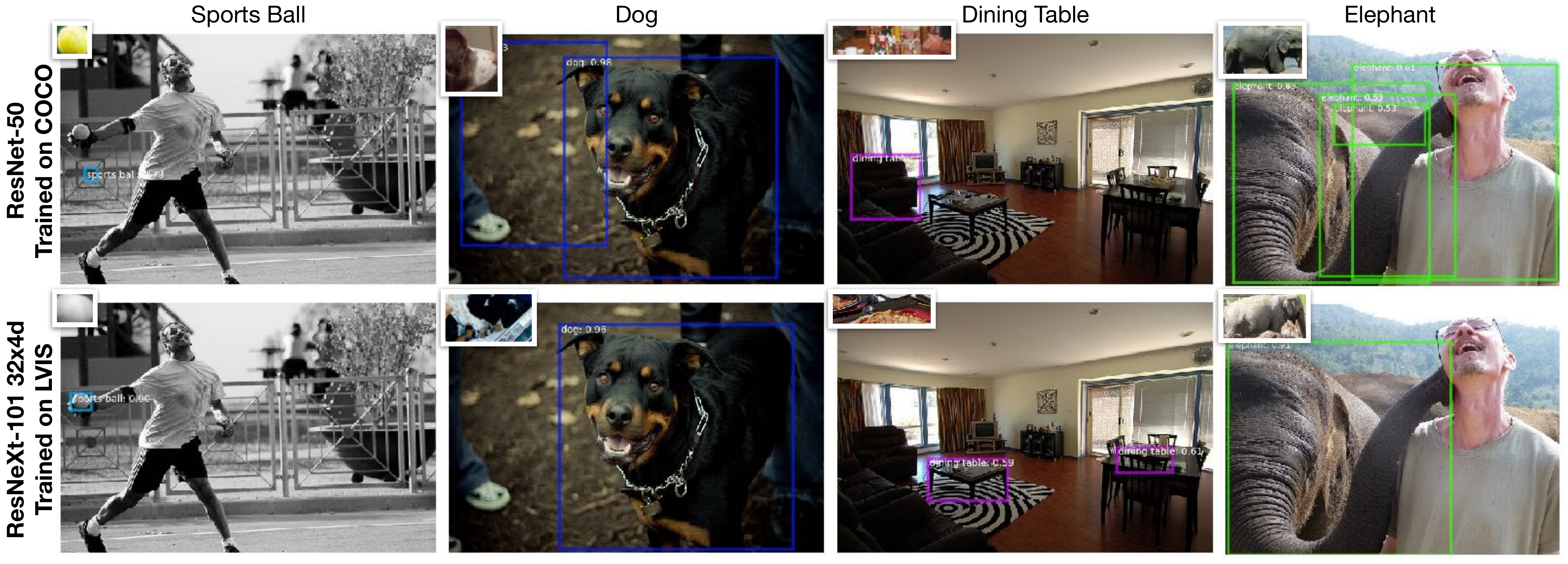}
    \caption{Predictions on COCO tend to be more accurate and cleaner when using a bigger backbone and training on LVIS. Especially on categories with more ambiguous references like sports ball or dining table the LVIS trained model is more precise. Additionally the ResNeXt backbone leads to "cleaner" results with less false positives.}
    \label{fig:examples_coco_comparison}
\end{figure}

\begin{table}[t]
    \centering
    \begin{footnotesize}
    \begin{tabular}{l|cc|cc|cc}
    \multicolumn{3}{c|}{} & \multicolumn{2}{c|}{1-shot} & \multicolumn{2}{c}{5-shot} \\
    Model & Backb. & Train Data & Train C. & Held-Out C. & Train C. & Held-Out C. \\
    \hline \hline 
    Siam. Mask R-CNN$^{*}$ & R50 & COCO & 37.6 & 16.3 & 41.3 & 18.5\\
    CoAE$^{**}$ & R50 & COCO & 40.9 & 22.0 & - & -\\
    AIT$^{***}$ & R50 & COCO & 47.5 & 24.3 & - & -\\
    \hline
    Siam. RetinaNet & R50 & COCO & 50.6 & 18.9 & 55.5 & 22.1\\
    Siam. Faster R-CNN & R50 & COCO & 49.7 & 22.8 & 54.9 & 27.6\\
    Siam. Mask R-CNN & R50 & COCO & 51.9 & 22.9 & 57.9 & 27.8\\
    Siam. Cascade R-CNN & R50 & COCO & 50.3 & 22.0 & 56.2 & 27.2\\
    Siam. Faster R-CNN & X101 32x4d & COCO & \textbf{56.4} & 23.5 & \textbf{61.9} & 28.6\\
    \hline
    Siam. Faster R-CNN & R50 & LVIS & 36.2 & 25.0 & 43.5 & 31.7\\
    Siam. Faster R-CNN  & X101 32x4d & LVIS & 42.5 & \textbf{27.4} & 50.3 & \textbf{34.8}\\
    \hline
    \end{tabular}
    \end{footnotesize}
    \caption{Performance (AP\textsuperscript{50}) on COCO can be improved by training on LVIS. Siamese Mask R-CNN and Siamese Cascade R-CNN are identical to Siamese Faster R-CNN except for an additional mask head or cascaded bbox heads. (*\cite{michaelis2018instance}, ** \cite{Hsieh2019coae}, *** \cite{chen2021ait})}
    \label{table:coco_sota}
\end{table}

Using the insights from above, we now demonstrate state-of-the-art one-shot detection performance on COCO by training on a large number of categories. We use LVIS and create four splits which leave out all categories that have a correspondence in the respective COCO split. As LVIS is a re-annotation of COCO, this means that we expand the categories in the training set while training on the same set of images.
Training with the more diverse LVIS annotations leads to a noticeable performance improvement from 22.8 to $25.0\,\% \text{AP}^{50}$, which can be improved even further to $27.4\,\%\text{AP}^{50}$ by using the stronger ResNeXt-101 backbone, outperforming the previous best model by $5.4\,\%\text{AP}^{50}$ (\tabref{table:coco_sota}). In relative terms that means going from 45\% relative performance to 65\%, thus substantially outperforming the previous best method (55\% relative performance \cite{Hsieh2019coae}) both in absolute and relative terms. Visual inspection of the results (\figref{fig:examples_coco_comparison}) shows cleaner predictions with less false positives especially for difficult reference images.

\section{Discussion}
\label{subsec:discussion}
It has long been assumed and recently shown \cite{sbai2020fewshotdata, Jiang2020fewshotdata, Fan2020fsod} that training with more categories improves few-shot learning performance.
However the question whether this is due to better overall model performance or better generalization has not been answered so far.
Our results show that the underlying mechanism is an improvement in generalization from 45\% relative performance on COCO to 89\% on LVIS.
The effect is consistent for different detectors, backbone architectures and training schedules which suggests that the effect will hold for almost any model.
If this trend continues with more categories object detection that generalizes to any object is within reach.
This, however, does not mean that one-shot object detection is ``solved''. There are at least three important steps to take:

First and foremost the performance of example-based object detectors has to improve significantly. 
Our experiments outline a path forward, demonstrating that methods that profit general object detection transfer to novel categories when the generalization gap is closed. 
Secondly, we have to better understand the mechanisms that lead to the generalization gap. Our results indicate that one of the main reasons is a shortcut \cite{geirhos2020shortcut} - memorizing the training categories. That stronger models also perform better on novel categories with progressive closing of the gap is an indicator that the key issue was indeed overfitting. However more investigation will be required to determine which factors are important. Is it the sheer number of categories or is it their diversity, granularity, frequency? Or is the main factor semantic relationship as results from \cite{sbai2020fewshotdata} and \cite{Jiang2020fewshotdata} suggest?  
Finally we have to find a way to transfer this success to smaller datasets with less categories. While we achieve a new state-of-the-art on COCO the generalization gap there (69\%) is still larger than on LVIS (89\%).

\subsection{Future datasets should focus on the diversity of categories.}
\label{subsec:dataset_recommendations}

Our findings have important implications for the design of future datasets. For the goal of generalization a broader range of categories is helpful at any dataset size (\figref{fig:class_fraction}B: green curve above blue curve at any data fraction), while from a certain point onward more examples per category lead to diminishing returns (\figref{fig:class_fraction}B: green curve flattens out). At a time where few-shot and long-tail problems become more important in computer vision this suggest that future data collection and annotation efforts should focus more on a broad set of categories and less on the number of instances for each of those categories. 

An open question is, how broad datasets have to be.
Despite being a big step forward, training on LVIS still leaves a small generalization gap that widens when using stronger models. In other words: some amount of overfitting on the training categories remains. The good news is that we don't see a saturation (\figref{fig:class_fraction}B: dark blue curves still rise at the maximum number of categories) so further increasing the number of categories should reduce the remaining gap.

\subsection{The bigger picture}
\label{subsec:outlook}

Our insight that applying existing methods on larger and more diverse datasets can lead to unexpected capabilities is mirrored in other areas. This phenomenon has been observed time and again and was termed the ``unreasonable effectiveness of data'' \cite{halevy2009unreasonable, sun2017revisiting} or the ``bitter lesson'' \cite{sutton2019bitter}. It played a key role in the breakthrough of DNNs thanks to ImageNet \cite{imagenet, krizhevsky2012imagenet} as well as recent results on game-play \cite{berner2019dota} or language modelling \cite{brown2020gpt3}. 
Recently \cite{kolesnikov2019bigtransfer} and \cite{radford2021clip} achieve impressive results demonstrating strong performance at one-shot and zero-shot ImageNet classification. As in our study, simple methods (transfer learning in \cite{kolesnikov2019bigtransfer} and unsupervised image captioning in \cite{radford2021clip}) on large and diverse datasets led to results that are far better than what one would have expected: Achieving ResNet performance with ~10 \cite{kolesnikov2019bigtransfer} respectively zero \cite{radford2021clip} annotated samples per class in their case; 89\% relative performance on LVIS in our case. We hope that by building on this insight we can soon move from trying to solve few-shot learning towards using few-shot learning to solve other problems.

\newpage



\begin{footnotesize}

\bibliographystyle{plain}
\end{footnotesize}

\newpage
\appendix
\renewcommand\thefigure{\thesection.\arabic{figure}}    

\section{Appendix}
\setcounter{figure}{0}

\subsection{Additional few-shot results}

\begin{figure}[h]
    \centering
    \includegraphics[width=1.00\linewidth]{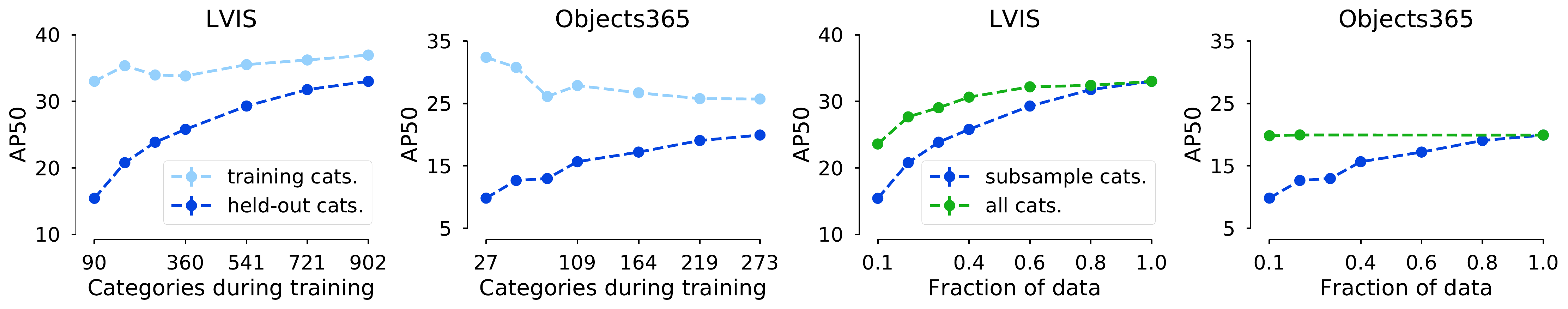}
    \caption{Performing the experiments in \figref{fig:class_fraction} with five reference images (five-shot) leads to no qualitative difference.
    }
    \vspace{0.3cm}
    \label{fig:class_fraction_5-shot}
\end{figure}

We provide five-shot results for the experiments in \figref{fig:class_fraction} in \figref{fig:class_fraction_5-shot}.

\end{document}